\documentclass[letterpaper]{article} 
\usepackage{aaai25}  
\usepackage{times}  
\usepackage{helvet}  
\usepackage{courier}  
\usepackage[hyphens]{url}  
\usepackage{graphicx} 

\usepackage{xcolor}
\usepackage{natbib}  
\usepackage{caption} 
\usepackage{algorithm}
\usepackage{algorithmic}
\usepackage{booktabs}
\usepackage{makecell}
\usepackage{multirow}
\usepackage{newfloat}
\usepackage{listings}
\DeclareCaptionStyle{ruled}{labelfont=normalfont,labelsep=colon,strut=off} 
\floatstyle{ruled}
\newfloat{listing}{tb}{lst}{}
\floatname{listing}{Listing}
\title{CrowdTraj: A Benchmark for Dense Crowd Trajectory Prediction in Realistic Crowded Environments}
\author{
    Antonius Bima Murti Wijaya,
    Paul Henderson,
    Marwa Mahmoud
}
\affiliations{
    \textsuperscript{\rm 1}School of Computing Science University of Glasgow\\

    University Avenue\\
    G12 8QQ, Glasgow, United Kingdom\\
    a.wijaya.1@research.gla.ac.uk, paul.henderson@glasgow.ac.uk, marwa.mahmoud@glasgow.ac.uk
}

\usepackage{bibentry}

\begin{document}

\maketitle

\begin{abstract}
In real-world applications, pedestrian trajectory prediction models rely on inputs from detection and tracking systems. Prior trajectory prediction benchmarks either contain relatively sparse pedestrian interactions, assume perfect tracking inputs, or rely on overhead viewpoints that minimize occlusion and perspective distortion, limiting evaluation in realistic dense-crowd scenarios.
We present CrowdTraj, a benchmark for pedestrian trajectory prediction in natural dense crowd scenes. Unlike previous datasets, CrowdTraj supports end-to-end evaluation from detection through tracking to trajectory prediction under severe occlusion in CCTV views. It also captures diverse, natural pedestrian behaviours, including abrupt directional changes rarely observed in existing benchmarks. CrowdTraj includes five diverse scenes, with an average of 1,146 unique pedestrians per scene, maximum frame-level densities ranging from 114 to 372 pedestrians, and over 3.2 million annotated head bounding boxes. CrowdTraj provides pixel and real-world coordinates via per-scene homography matrices for physically meaningful analysis.
Our experimental results show that tracking accuracy (IDF1) drops to 0.68–0.70 in the densest scenes, compared with approximately 0.90 in less crowded scenes. Trajectory prediction training also becomes substantially more computationally expensive in dense scenes, with training times increasing by up to 8 times. These findings show that CrowdTraj exposes limitations in current trajectory prediction pipelines that remain hidden on existing sparse-crowd benchmarks, particularly in robustness to tracking noise and computational scalability.
\end{abstract}

%

\section{Introduction}
Automated trajectory prediction systems rely on detection and tracking, yet nearly all existing benchmarks assume perfect trajectories and relatively sparse scenes. In a dense crowd scenario, full-body appearances are not always available due to heavy occlusion, especially in CCTV views, which are widely deployed in many areas. The current annotation that works in crowds usually relies on head detection \cite{Sundararaman2021,Cao2017, Sun2025, Badauraudine2025}. On the other hand, relying on head labels reduces the amount of information available to the pedestrian for tracking, which could lead to further performance degradation across all tasks. Working with the crowd means that the challenge also lies in the number of objects to be processed, which could bring issues in efficiency.

Widely used trajectory prediction datasets have fewer crowds and are collected in less-occluded environments, with the ETH-UCY and SDD datasets having an average density of fewer than 15 pedestrians per frame.(see Table \ref{tab:dataset_stats}) \cite{Pellegrini2009, Lerner2007, Robicquet2016}, and current trajectory prediction algorithms have shown promising performance on these datasets. Some publicly available datasets work on crowd scenarios, which are Head Tracking 21 (HT-21)\cite{Sundararaman2021} and the Jüliech dataset \cite{Cao2017}. The HT-21 dataset includes a real-world scene of tracking and trajectory prediction; however, they do not provide a long sequence for trajectory prediction. The Jüliech dataset provides a long sequence of a setup scene instead of in the wild and taken from very high altitude or overhead views, which reduces the possibility of occlusion and trajectory outliers. 

CrowdTraj enables realistic end-to-end evaluation from detection through tracking to trajectory forecasting under severe occlusion and noise with a large number of pedestrians. Our dataset includes real-world scenes of pedestrians walking in 5 scenes, including train stations, busy pedestrian walks, and crowd events. Where people are eventually occluded by other persons or other obstacles in a crowd, which is challenging for the whole trajectory prediction task. Our dataset offers 5729 total unique pedestrians, with a peak of 372 pedestrians in a single frame, and each scene lasts approximately 4 minutes. To the best of what we know, CrowdTraj is the only dataset that has CCTV views with high density, varying trajectory patterns, and long sequences; it can facilitate all of the automatic dense crowd trajectory prediction stages.

We evaluated several state-of-the-art tracking algorithms and trajectory prediction methods using our dataset, evaluating them on 24 historical and 36 future frames. Based on the experiment result, our dataset reveals a substantial gap in processing trajectory prediction and tracking performance.

Our main contributions can be summarized as follows: 
\begin{itemize}
\item We introduce CrowdTraj, a dense crowd dataset for end-to-end trajectory prediction that spans detection, tracking, and forecasting. Our real-world coordinates via homography metrics enable accurate distance-based calculations, such as velocity, acceleration, and interaction radius.

\item We benchmark several state-of-the-art tracking and trajectory prediction methods on CrowdTraj, showing that performance degrades substantially under high density and severe occlusion, both in terms of accuracy and computational efficiency, relative to their reported performance on existing sparse-crowd benchmarks.

\item We show that CrowdTraj exposes failure modes in modern end-to-end trajectory prediction pipelines that remain largely hidden on existing sparse-crowd benchmarks, highlighting important challenges in robustness to tracking noise and computational scalability.
\end{itemize}

\begin{table*}[t]
\centering
\caption{Dataset Statistics Comparison between the common trajectory dataset and the dense crowd dataset. The top rows show summary statistics for every dataset per scene, and the average max density column indicates the average maximum number of pedestrians that appeared on a frame among all scenes. The Avg Traj length has units in pixels. The Bottom rows show scene statistics of our dataset, where the Max Density indicates the maximum number of pedestrians on a frame in the scene.}
\label{tab:dataset_stats}
\begin{tabular}{l l r r r r r r r  r}
\toprule
Dataset & 
\makecell{Axis} & 
\makecell{Unique\\Peds} & 
\makecell{Avg.\\Length (s)} & 
\makecell{Total\\Frames} & 
Boxes & 
\makecell{Avg.\\Density} & 
\makecell{Avg Max\\Density} & 
\makecell{Frame\\Rate(fps)} \\
\midrule
ETH-UCY   & Real  & 275.6  & 372.1  & 9.3K   & 9.3K   & 14.9  & 49.1& 25 \\
SDD       & Pixel  & 654.0  & \bfseries 1,646.2 & \bfseries 49.4K  & 742K & 11.9  & 17.4  & \bfseries 30 \\
CCHead    & Real	& 165.64 & 169.27 	& 4.2K 	&181.8K &95.18 	& 67.2 & 25\\ 
HT-21   & Pixel & 688.0  & 57.4   & 1.4K   & 297.1K & \bfseries 176.3 & 257.9 & 25 \\
Jülich   & Pixel & 637.6  & 250.1  & 2.5K   & 368K & 144.4 & 174.5  & 10 \\
CrowdTraj(Ours) & Real & \bfseries 1,145.8 & 248.6  & 4.9K   & \bfseries 642.5K & 149.5 & \bfseries 312.3  & 20 \\
\midrule
Duri Morning (DM)    & Real & 1,112 & 241.1 & 4.8K & 446.7K  & 92.63  & 115   & 20 \\
Duri Evening (DE)  & Real & 1,181 & 250.4 & 5.0K & 383.6K  & 76.61  & 114 & 20 \\
Vredeburg (VB)      & Real & 934   & 250.3 & 5.0K & 622.0K & 124.28 & 173& 20 \\
Duri Platform1 (P1)  & Real & 1,275 & 250.8 & 5.0K & 711.6K  & 141.86 & 182  & 20 \\
Paisley Festival (PA)      & Real & 1,227 & 250.8 & 5.0K & 1,048,6K & 312.27 & 372  & 20 \\
\bottomrule
\end{tabular}
\end{table*}
\section{Related Works}
\subsection{Pedestrian Trajectory Prediction dataset}
Trajectory prediction is a comprehensive task in computer vision because it involves several tasks, including detection and tracking. Several datasets are: ETH\cite{Pellegrini2009}, UCY\cite{Lerner2007}, SDD \cite{Robicquet2016}, CChead \cite{Sun2025}, built to support this trajectory prediction task, which are commonly used in trajectory prediction algorithm research \cite{Salzmann2020, Ivanovic2019, XuVAE2022, Zhou2022, Lee2024, Fu2025}. However, those datasets do not contain crowds in their scenes except for the rooftop scene on CChead with a maximum of 115 pedestrians in a frame, and it still only contains 96 seconds, both for training and testing.

There are several other datasets that have dense crowd scenes, which are the Jülich dataset\cite{Cao2017} and the Head Tracking 21 (HT-21) \cite{Sundararaman2021} dataset. Both of them use the head bounding box instead of the full body appearance due to a highly occluded situation and were captured from two different viewpoints: HT-21 was taken from a CCTV view, which offers a challenge in tracking due to high occlusions. Therefore, HT-21 is challenging to work on object detection, tracking, and trajectory prediction tasks simultaneously. The Jülich \cite{Cao2017} dataset was taken from an overhead view, which brings a clear view with less occlusion. The Jülich dataset offers a longer duration, which could be more than 5 minutes. The Jülich dataset was created in setup scenes, where people are instructed to follow the direction of the area. Therefore, the Jülich dataset can comprehensively challenge the trajectory prediction task only, instead of tracking and detection. Even though the Jülich dataset provides the site maps with real-world scale, they do not provide the camera intrinsic parameters, which may lead to inaccurate real world coordinate.

Currently, two emerging datasets have been proposed to challenge the trajectory prediction task, which are Gigatraj\cite{Lin2024}, which is based on the Panda dataset, and the Lyon Crowd dataset, which was taken at the Festival of Light in 2022\cite{Dufour2025}. Both of them were taken with a sophisticated camera to cover a massive area over a long duration. Although the Lyon dataset captures dense crowds, only a random sample of individuals is tracked in the wide-area views, rather than the full set of pedestrians in the scene; exhaustive tracking is limited to a small sub-region over a short duration. This makes it unsuitable for trajectory prediction, which typically requires complete neighborhood information to model pedestrian interactions. On the other hand, Gigatraj shows a great sequence of natural pedestrians in a broad view. Even though the pedestrian data is large due to the broad view, they are not a dense crowd but more likely sparse.

\subsection{Detection-Tracking to Trajectory Predictions}The natural use of trajectory prediction is to feed the input from tracking results, which also elaborates the detection task. YOLO has become one of the leading online detectors, which currently reaches versions v11 \cite{yolo11_ultralytics},v12\cite{yolo2023}, and v26\cite{yolo26_ultralytics}. Several tracking algorithms are usually used to track pedestrians based on this online detector\cite{Cao2023, Zhang2022, Aharon2022, WangSmile2024, Erregue2025}. Other trackers specifically track the head of pedestrians and use their own tracker in an offline manner, such as Headhunter-T\cite{Sundararaman2021} and Multi-source Data Fusion Network (MDFN)\cite{Fu2025}. Those trackers do not perfectly track the object without any issues, such as ID-switch and fragmentation, which will become noise for trajectory prediction input \cite{chib2024}. As a result, trajectory prediction is usually treated separately from tracking. 
The basis of trajectory prediction is not only using the agent's previous input (usually 8 observed steps and 12 future steps) but also neighbourhood data \cite{Alahi2016, Salzmann2020, Xugroupnet2022, XuVAE2022, Zhou2022, Lee2024, Fu2025, Chen2025}, such as other pedestrians or other objects that may affect the pedestrian trajectories. Some of the research also includes a map of the area and real-world coordinates as multimodality to improve the accuracy \cite{wongsc2024, Fu2025}. This entanglement with neighborhood data makes trajectory prediction in crowds especially challenging, due to both heavy computation and noisy input patterns. CrowdTraj contains crowd data that could challenge the effectiveness of the trajectory prediction algorithm and the accuracy as well. CrowdTraj also has scenes with various trajectory patterns, like a sharp turn movement that become a challenge for longer trajectory prediction, where some researchers have already discussed long-term prediction \cite{Lin2024}.

\section{CrowdTraj Benchmark}
CrowdTraj labels follow the MOT(Multiple Object Tracking)\cite{Rudenko2020,Sundararaman2021} format, which contains frame numbers, pedestrian IDs, top-left x, top-left y, width, and height, and the image data uses 4K format with dimensions of 3840x2160. 
\subsection{Data Collection}
\begin{figure}[t]
\centering
\includegraphics[width=0.45\textwidth]{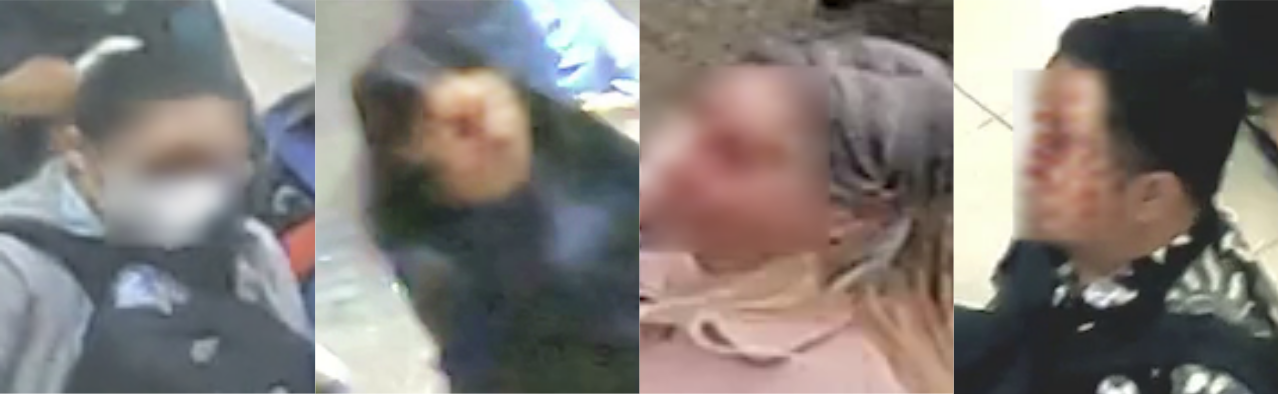} 
\caption{The DartBlur\cite{Jiang2023} implementation on our dataset to blur the detected pedestrian faces}
\label{fig:dart}
\end{figure}

Our dataset is taken from several places in Indonesia and Scotland. The places are located at the Duri Station Hub in Indonesia, a public pedestrian walkway in front of Fort Vredeburg (VB), and at the Paisley Food and Drink Festival (PA). At Duri Station, we took footage during morning scenes(DM), evening scenes(DE), and at the Platform 1 entrance(P1). We have already gained permission from all venue managers, including PT KAI(Kereta Api Indonesia) Region 1 as the venue manager and the Paisley Festival Organizer. The data is published under the Creative Commons Attribution-NonCommercial ShareAlike 4.0 license. Our dataset already anonymizes the faces of the pedestrians to protect individual privacy using DartBlur\cite{Jiang2023} on identified faces; see figure \ref{fig:dart}. We also placed a sign with a poster stating that a video is being recorded in the area.
We used a GoPro camera taken at certain heights(4-5 meters), depending on the area, to capture the scenes. We recorded it for around 5 to 6 minutes for each scene and cut it down to around 5000 frames based on the crowd density in the frames. 
\subsection{Data Annotation}
 We have 7 people working as annotators, where 3 work on Paisley festival scenes. Our annotation methods are semi-automatic, which means we did automatic tracking and then refined the result manually. First, we tagged 100 frames taken randomly with a head bounding box for each scene. Then, train the head dataset for YOLOv11 detection, and the head detection models are then used to run the tracking algorithm across all scenes. They looked at the tracking mistakes, such as identity switch, track fragmentation, and loss of track, and then manually fixed them. Our annotation result does not include when the object is not visible or occluded; it appeared as no track for the pedestrian in several frames, which is due to the high rate of occlusion, where the pedestrians are highly likely to disappear from the videos. A pedestrian is treated as the same identity if they can be visually re-identified, especially if this happens in less than 100 frames. If it is more than 100 frames and the pedestrian can not be recognized, it is considered a new track. 
\subsection{Quality Control}
\begin{table}[t]
\centering
\caption{Ambiguity rate for each scene in CrowdTraj.}
\label{tab:scene_ambiguity1}
\begin{tabular}{lccccc}
\toprule
\textbf{Scene} & \textbf{DM} & \textbf{DE} & \textbf{VB} & \textbf{P1} & \textbf{PA} \\\\
\midrule
Ambiguity (\%) & 0.85 & 1.31 & 0.99 & 2.52 & 1.45\\ \bottomrule
\end{tabular}
\end{table}
All the annotators reviewed others anotators works, they revisited frames from the beginning to check and refine the labels. However, an ambiguity arose in determining the head detection in low visibility due to the distance from the camera, motion blur, and head color similarity at which a head could still be reliably identified. Therefore, we also measure this by comparing manual head detection of 60-100 random frames from cross-annotators to the ground truth. The ambiguity is calculated by finding the difference in the number of pedestrians on selected frames, and we found that our average ambiguity score across all scenes is 1.42\% per frame. Based on Table \ref{tab:scene_ambiguity1} P1 has the highest ambiguity score since the head mostly blended with the shadows in a crowd.

\subsection{Real World Coordinate and Distortion}
We have taken manual real-world coordinate measurements for the Duri station (DM, DE, P1) scenes and the Fort Vredeburg scene(VB). For the Paisley festival scene(PA), we took the coordinates from Google Maps since it has better accuracy than manual measurements because it is hard to find and measure manual anchors on the sites. The trajectory map with real-world coordinates is transformed from pixel coordinates to the perspective of the real world. For camera calibration, since our video was taken with a super wide camera, we also give our best camera calibration settings with cv2 calibration with the fisheye class to accommodate the expansion at the corner. We use the remapping technique to undistort the image and apply the ray plane intersection method to calculate the homography matrix.

\subsection{Data Statistics and the Scenes}

\newcommand{\topimg}[1]{%
  \begin{minipage}[t]{\linewidth}
  \vspace{0pt}
  \includegraphics[width=2.9cm]{#1}
  \end{minipage}
}
\renewcommand{\arraystretch}{1.4}

\begin{table*}[h!]
\centering

\begin{tabular}{
>{\raggedright\arraybackslash}p{2cm}@{\hspace{3pt}}
p{3cm}@{\hspace{1pt}}
p{3cm}@{\hspace{1pt}}
p{3cm}@{\hspace{1pt}}
p{3cm}@{\hspace{1pt}}
p{3cm}
}
\hline

\textbf{Scenes} & DM & DE & VB & P1 & PA \\
\hline

\vspace{0pt} Sample&
\topimg{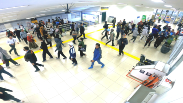} &
\topimg{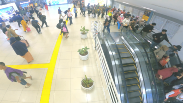} &
\topimg{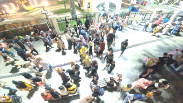} &
\topimg{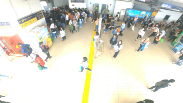} &
\topimg{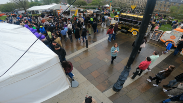} \\

\vspace{0pt} Pixel Unit Pattern  &
\topimg{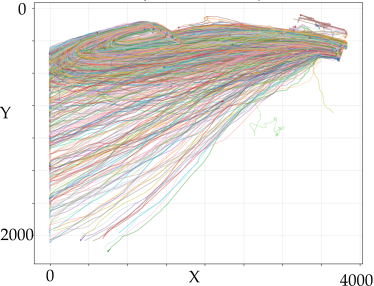} &
\topimg{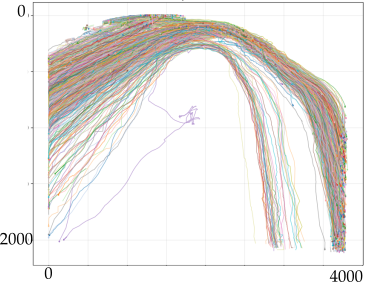} &
\topimg{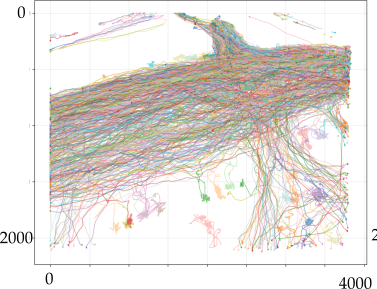} &
\topimg{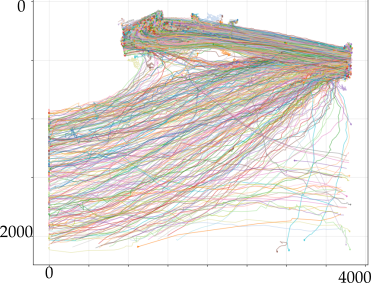} &
\topimg{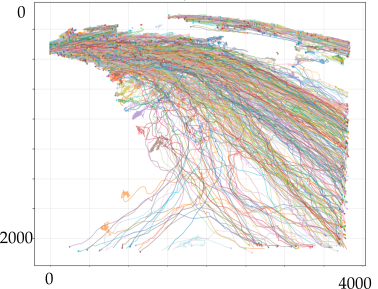} \\

\vspace{0pt} World Unit Pattern (cm) &
\topimg{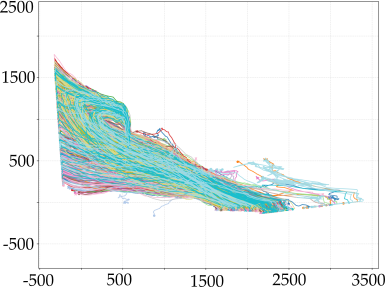} & \topimg{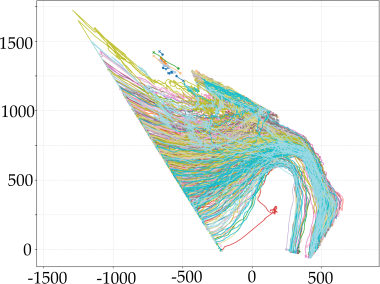} & \topimg{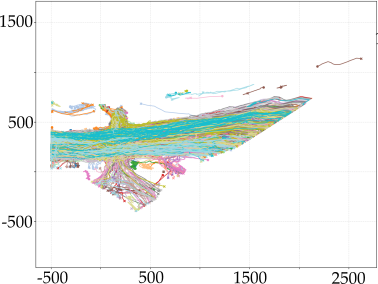} & \topimg{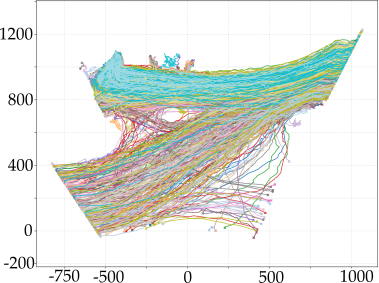} & \topimg{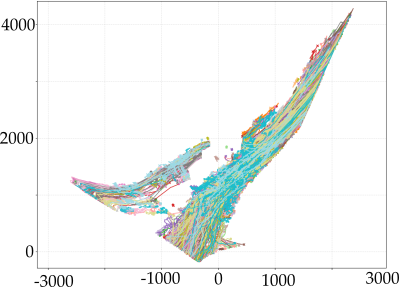} \\

\vspace{0pt} Pedestrian Number(X:peds number, Y: Frames)&
\topimg{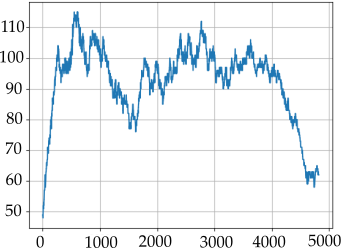} &
\topimg{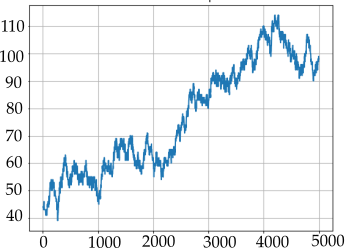} &
\topimg{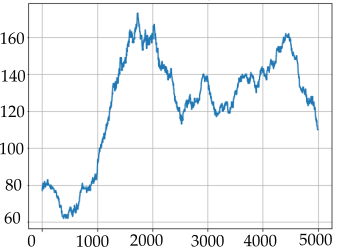} &
\topimg{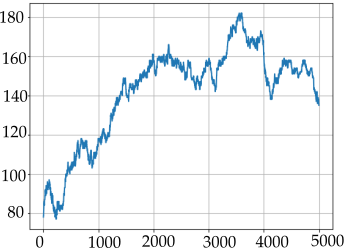} &
\topimg{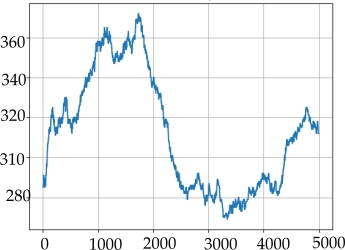} \\

Mean/Median Velocity (m/s)&
1.67/1.62 & 1.06/0.94 & 0.68/0.62 &0.70/0.63 &0.71/0.59 \\

Mean/Median distance/person (m)&
25.50/28.09 & 13.52/14.69 & 16.37/15.79 &13.54/15.83 & 25.04/21.91 \\

\hline
\end{tabular}

\caption{The table shows both pixel pattern and a real-world coordinate translation pattern. The table also shows the crowd's behaviour through the pedestrian number, average velocity, and travel distance across the scenes.}
\label{tab:dataset_pattern}
\end{table*}

Compared to other datasets, as seen in Table \ref{tab:dataset_stats}, our dataset has the largest average number of pedestrians, bounding boxes, and max density across all frames and scenes, demonstrating that CrowdTraj captures substantially higher crowd density than existing benchmarks. Compared to non-crowd pedestrian scenes (ETH-UCY, SDD, CCHead), SDD contains more total minutes and frames; however, despite its shorter total duration, CrowdTraj contains more unique pedestrians than SDD. If we specifically compare with the other crowd datasets (MOT-21 and Jülich), our dataset has both real-world coordinates and pixel coordinates; therefore, it can be used for further location-based analysis and its derivatives. 

Beyond aggregate density, CrowdTraj captures a wide range of pedestrian motion regimes across scenes, from goal-directed, time-pressured movement to leisurely, exploratory wandering a diversity as detailed below.

DM and DE scenes have the fewest pedestrians among the others; pedestrians in these scenes exhibit sharp directional turns toward the other platform(see Table \ref{tab:dataset_pattern}). The DM scene has the highest velocity since it was taken when people are hurrying to work. On the other hand, the evening scene was taken during peak hour while people were going back home. Therefore, some of them are not in a rush, as in the morning.

At P1 scenes, a bottleneck occurs toward the Platform 1 entrance. Pedestrian density gradually increases over the course of the scene due to the bottleneck(see table \ref{tab:dataset_pattern} on the pedestrian number row). This scene was recorded in the morning, when pedestrians are initially rushing but slow down due to the bottleneck.

Meanwhile, the Paisley festival (PA) and the Fort Vredeburg(VB) scenes are both in more enjoyable moments. The Vredeburg scene is centered on a three-way junction, where pedestrians are free to choose their direction after exiting the gate. Meanwhile, pedestrian movement in the Paisley scene is largely bidirectional; however, it has the most pedestrians among all of the scenes, reaching 372 pedestrians in a frame. As seen in Table \ref{tab:dataset_pattern}, the Paisley trajectory is rising in the beginning and decreases in the middle. t also contains two independent sub-scenes, one along the main path and one along a smaller side path that do not interact with each other.

\section{Algorithm Analysis}
\begin{table*}[t]
\centering
\caption{Comparison of Tracking Methods based on YOLOv26(ByteTrack,BOT-SORT SMILETrack) and YOLOv11 JDE across Multiple Scenes}
\label{tab:tracking_results_final}
\small
\begin{tabular}{@{}llcccccccc@{}}
\toprule
\textbf{Scene} & \textbf{Method} & \textbf{TP} ($\uparrow$) & \textbf{FP} ($\downarrow$) & \textbf{FN} ($\downarrow$) & \textbf{IDSW} ($\downarrow$) & \textbf{MOTA} ($\uparrow$) & \textbf{Prec.} ($\uparrow$) & \textbf{Rec.} ($\uparrow$) & \textbf{IDF1} ($\uparrow$) \\ \midrule

\multirow{4}{*}{DM} 
 & ByteTrack & 381,234 & 16,578 & 65,440 & 3,803 & 0.81 & 0.96 & 0.85 & 0.90 \\
 & BoT-SORT & \textbf{383,069} & 17,434 & \textbf{63,605} & 4,180 & \textbf{0.81} & 0.96 & \textbf{0.86} & \textbf{0.90} \\
 & SMILETrack & 372,433 & \textbf{10,792} & 74,241 & \textbf{2,548} & 0.80 & \textbf{0.97} & 0.83 & 0.89 \\
 & JDE-YOLOv11 & 380,761 & 16,430 & 65,913 & 5,453 & 0.80 & 0.96 & 0.85 & 0.90 \\ \midrule

\multirow{4}{*}{DE} 
 & ByteTrack & 295,971 & 17,577 & 87,594 & 3,324 & 0.71 & 0.94 & 0.77 & 0.85 \\
 & BoT-SORT & \textbf{299,958} & 18,327 & \textbf{83,607} & 4,150 & \textbf{0.72} & 0.94 & \textbf{0.78} & \textbf{0.86} \\
 & SMILETrack & 285,847 & \textbf{11,910} & 97,718 & \textbf{1,973} & 0.71 & \textbf{0.96} & 0.74 & 0.84 \\
 & JDE-YOLOv11 & 283,120 & 16,271 & 100,445 & 5,284 & 0.68 & 0.95 & 0.74 & 0.83 \\ \midrule

\multirow{4}{*}{VB} 
 & ByteTrack & 441,621 & 53,509 & 180,415 & 3,747 & 0.62 & 0.89 & 0.71 & 0.79 \\
 & BoT-SORT & \textbf{443,623} & 55,488 & \textbf{178,413} & 4,523 & 0.61 & 0.89 & \textbf{0.71} & \textbf{0.79} \\
 & SMILETrack & 431,866 & 38,035 & 190,170 & \textbf{2,270} & \textbf{0.63} & \textbf{0.92} & 0.69 & 0.79 \\
 & JDE-YOLOv11 & 366,337 & \textbf{36,932} & 255,699 & 8,705 & 0.52 & 0.91 & 0.59 & 0.71 \\ \midrule

\multirow{4}{*}{P1} 
 & ByteTrack & 438,081 & 134,616 & 273,506 & 6,921 & 0.41 & 0.76 & 0.61 & 0.68 \\
 & BoT-SORT & 441,005 & 141,401 & 270,582 & 8,815 & 0.41 & 0.75 & 0.62 & 0.68 \\
 & SMILETrack & 408,725 & \textbf{102,265} & 302,862 & \textbf{2,865} & 0.45 & \textbf{0.80} & 0.57 & 0.67 \\
 & JDE-YOLOv11 & \textbf{452,704} & 124,694 & \textbf{258,883} & 9,813 & \textbf{0.45} & 0.78 & \textbf{0.64} & \textbf{0.70} \\ \midrule

\multirow{4}{*}{PA} 
 & ByteTrack & 1,121,268 & 119,134 & 445,103 & 19,116 & 0.63 & 0.90 & 0.72 & 0.80 \\
 & BoT-SORT & 1,129,458 & 132,202 & 436,913 & 23,677 & 0.62 & 0.89 & 0.72 & 0.80 \\
 & SMILETrack & 994,786 & \textbf{31,072} & 571,585 & \textbf{5,639} & 0.61 & \textbf{0.97} & 0.63 & 0.77 \\
 & JDE-YOLOv11 & \textbf{1,151,341} & 78,964 & \textbf{415,030} & 14,238 & \textbf{0.68} & 0.93 & \textbf{0.74} & \textbf{0.82} \\ \bottomrule
\end{tabular}
\end{table*}

\begin{table*}[ht]
\centering
\setlength{\tabcolsep}{2pt}
\caption{Performance comparison of SocialVAE*, MART, MoFlow, and DSTIGCN across all scenes. MART-SMILE reflects MART's performance using SMILETrack tracking output as input, evaluated against ground-truth trajectories. MART-World reports MART's performance on real-world coordinates.}
\label{tab:trajectory_results}
\begin{tabular}{lcccccccccc}
\toprule
& \multicolumn{2}{c}{DM} 
& \multicolumn{2}{c}{DE}
& \multicolumn{2}{c}{VB}
& \multicolumn{2}{c}{P1}
& \multicolumn{2}{c}{PA} \\
\cmidrule(lr){2-3} \cmidrule(lr){4-5} \cmidrule(lr){6-7} \cmidrule(lr){8-9} \cmidrule(lr){10-11}
Method 
& ADE/FDE & Time 
& ADE/FDE & Time
& ADE/FDE & Time 
& ADE/FDE & Time 
& ADE/FDE & Time \\
\midrule
SocialVAE* 
& 9.18/11.74 & 824 
& 10.46/10.61 & 635 
& 10.63/11.69 & 945 
& 5.12/6.61 & 1135 
& 5.60/7.06 & \textbf{1950} \\
MoFlow
& 16.00/27.42 & 524 
& 15.51/20.90 & 701 
& 16.55/24.55 & 746 
& 14.24/17.50 & 1290 
& 8.15/11.88 & 2211 \\
DSTIGCN
& 13.00/24.42 & 916
& 13.33/20.97 & 512 
& 13.71/22.24 & 2503
& 7.27/11.60 & 1837
& 6.16/9.81 & 3703 \\
MART 
& \textbf{6.47}/\textbf{9.93} & \textbf{337} 
& \textbf{9.08}/\textbf{11.09} & \textbf{331} 
& \textbf{9.43}/\textbf{12.95} & \textbf{461}
& \textbf{4.59}/\textbf{6.43} & \textbf{516} 
& \textbf{4.00}/\textbf{5.29} & 2456 \\
\midrule
MART-SMILE
& 6.75/10.49 & 397
& 9.21/11.35 & 276 
& 9.76/13.33 & 368 
& 4.79/6.65 & 694  
& 4.18/5.57 & 1532  \\ 
MART-World
& 6.56/10.84 & 697
& 4.93/6.29 & 786 
& 5.69/7.45 & 1146 
& 4.08/5.78 & 627  
& 5.38/6.67 & 3155 \\
\bottomrule
\end{tabular}
\end{table*}
In this section, we analyse CrowdTraj across the tasks required for automatic trajectory prediction: tracking by detection and trajectory prediction. Since tracking depends on detection output, we combine these two into a single detection-and-tracking experiment. The trajectory prediction is evaluated on ground-truth tracks, not on tracker outputs. We then extend the evaluation by feeding the best-performing tracker's output into the best-performing trajectory prediction model, to measure performance under realistic, noisy end-to-end conditions.
\subsection{Baseline}
In this study, we employ widely adopted real-time detection and tracking approaches, including YOLOv11 and YOLOv26 as the detector, combined with several widely used trackers that has effort to tackle occlusion and reidentification problems such as ByteTrack, BoT-SORT, SMILETrack, and JDE. 

A key characteristic of ByteTrack is its strategy of leveraging both high- and low-confidence detections, enabling it to recover missed associations and maintain track continuity under challenging conditions, such as occlusion \cite{Zhang2022}. In contrast, BoT-SORT extends motion-based tracking by incorporating appearance cues extracted from bounding boxes. By combining motion prediction with appearance-based re-identification features, BoT-SORT improves identity preservation and robustness in crowded or dynamic scenes\cite{Aharon2022}. JDE (Joint Detection and Embedding) adopts a unified framework in which detection and re-identification are learned simultaneously. Through self-supervised ReID feature learning, JDE produces embeddings that facilitate consistent identity matching across frames without requiring a separate feature extractor\cite{Erregue2025}. Finally, SMILETrack enhances tracking robustness by using learned similarity. Through the Siamese Network, SMILETrack calculates the appearance similarity between two different object identifiers to determine whether both belong to the same object or not to reduce the ID-Switch problem\cite{WangSmile2024}.

In the trajectory prediction case, we compare four different algorithmic approaches. SocialVAE \cite{XuVAE2022} constructs a scene-level neighbourhood graph before passing it to the model, which requires substantial memory to build, so it requires huge memory to develop the neighbourhood graph. However, the original socialVAE requires an agent to connect to all neighbours on the preparation stage, which consumes vast memory; therefore, we update the socialVAE to only accept a maximum of 20 closest neighbours at this stage to reduce memory consumption and improve generalization. MART \cite{Lee2024}  does not construct the scene graph at the input level; instead, it builds a hypergraph for each batch during training. Therefore, MART can manage the computation efficiently and process it with a Transformer. We also compare the trajectory prediction with MoFlow \cite{Fu2025}, which is the current methods that focus on providing diverse and plausible outputs with flow matching loss. The last is DSTIGCN \cite{Chen2025}, which also uses an attention mechanism for pedestrian interaction like MART and further processes it with a Graph Convolution Network.

\subsection{Algorithm Evaluation Metrics}
Since we conduct two distinct experiments (Tracking by Detection and Trajectory prediction), we also have different metrics for each task, which are widely used for these tasks.
For tracking, we use: True Positive(TP), False Positive(FP), False Negative(FN), Identity Switch(IDSW), Multiple Object Tracking Accuracy (MOTA), Precision (Prec.), Recall(Rec),  and identity preservation capability of a multi-object(IDF1)

Following common practice in trajectory prediction, we report minADE (Average Displacement Error)and minFDE(Final Displacement Error), where the prediction with the smallest error relative to the ground truth is selected for each agent. 
\subsection{Experiments Details}

\begin{table}[t]
\centering
\footnotesize
\setlength{\tabcolsep}{4pt}
\caption{Extended experiment on MART and SocialVAE for the Generalization Ability on real-world unit accuracy. Model means the model that is trained on the concurrent data, and the data means the data source for the evaluation. Our SocialVAE* shows better performance on real-world coordinate prediction and also for the generalization ability.}
\label{tab:forget_compare}
\begin{tabular}{llccccc}
\toprule
 &  & \multicolumn{5}{c}{\textbf{Model}} \\
\cmidrule(lr){3-7}
Data & Method & DM & DE & VB & P1 & PA \\
\midrule
\multirow{2}{*}{DM}
& MART      & \textcolor{blue}{0.070} & 0.149 & 0.138 & 0.152 & 0.140 \\
& SocialVAE* & \textbf{\textcolor{blue}{0.056}} & \textbf{0.111} & \textbf{0.084} & \textbf{0.085} & \textbf{0.091} \\
\midrule
\multirow{2}{*}{DE}
& MART      & 0.085 & \textcolor{blue}{0.049} & 0.074 & 0.077 & 0.067 \\
& SocialVAE* & \textbf{0.073} & \textbf{\textcolor{blue}{0.047}} & \textbf{0.056} & \textbf{0.056} & \textbf{0.059} \\
\midrule
\multirow{2}{*}{VB}
& MART      & \textbf{0.049} & 0.087 & \textcolor{blue}{0.057} & 0.075 & 0.071 \\
& SocialVAE* & 0.072 & \textbf{0.076} & \textbf{\textcolor{blue}{0.056}} & \textbf{0.072} & \textbf{0.063} \\
\midrule

\multirow{2}{*}{P1}
& MART      & 0.085 & 0.074 & 0.049 & \textcolor{blue}{0.041} & 0.050 \\
& SocialVAE* & \textbf{0.032} & \textbf{0.031} & \textbf{0.031} & \textbf{\textcolor{blue}{0.024}} & \textbf{0.032} \\
\midrule

\multirow{2}{*}{PA}
& MART      & 0.085 & 0.083 & 0.081 & 0.082 & \textbf{\textcolor{blue}{0.054}} \\
& SocialVAE* & \textbf{0.075} & \textbf{0.070} & \textbf{0.066} & \textbf{0.066} & \textcolor{blue}{0.059} \\
\bottomrule
\end{tabular}
\end{table}
In this paper, we present an experiment on tracking by detection and trajectory prediction, in which both operate independently to achieve the maximum performance boundary for trajectory prediction.

On tracking experiments, we use the first 1000 frames for training and the next 200 frames for validation in every scene. We sample every 5 frames for both the training and validation sets for YOLOv26 and YOLOv11.

On trajectory prediction, we use 3 splits for training, validating, and evaluating. The training set was taken based on the pedestrian number graph shown in Table \ref{tab:dataset_pattern}, which makes sure the training contains most of the trajectories. The training set took a longer horizon than common trajectory predictors; we took 24 observations and 32 future steps, resulting in observation and prediction horizons approximately three times longer than the standard 8/12 setting. Our 20 neighbourhood limitation on SocialVAE helps manage memory consumption since it requires more than 200GB of memory for the full neighbourhood at the beginning. However, for other methods, we can consider using the whole neighbour graph since it divides the neighbour data into batches. We calculate the training time based on the last best-performing validation model saved. None of the experiments used multimodal features, even where supported by the original baseline implementation.

\section{Results}
The experimental results indicate that Duri Morning and Duri Evening represent the least challenging tracking tasks, primarily due to lower pedestrian density compared to the Paisley and Duri Platform 1 scenarios. In these less crowded environments, BoT-SORT-YOLOv26 demonstrated superior performance, achieving a MOTA of approximately 0.81 and an IDF1 of 0.90. While other trackers produced slightly lower metrics, the performance gap between models in these scenarios remained marginal.

The complexity increases significantly in the Paisley and Duri Platform 1 scenes due to high occlusion levels. In the Paisley scene, tracking is hindered in the background areas where individuals frequently disappear into the crowd. Conversely, Duri Platform 1 presents a unique challenge at the entrance, where a "bottleneck" effect and low visibility lead to the lowest recorded MOTA and IDF1 scores across all tasks. On these high-difficulty tasks, SMILETrack-YOLOv26 and JDE-YOLOv11 exhibited greater robustness. The performance gain is limited by the fact that the models are configured for head-only detection rather than full-body tracking.

Trajectory prediction yielded different results; MART achieved the lowest error and the shortest training time for 4 of 5 scenes. SocialVAE*(our modified 20-neighbor variant) showed in second place for accuracy. The other methods, such as DSTIGCN and MoFlow, did not achieve comparable accuracy and were highly prone to gradient explosion. We had to reduce the batch and the learning rate and apply epsilon(a very small floating-point value) to the loss function to prevent invalid values.

Although MART trained fastest on most scenes, this trend reversed on the Paisley Festival scene, where its training time increased substantially, even exceeding SocialVAE*'s. This is because, in this very dense scenario, our neighbourhood limitation in SocialVAE keeps the number of neighbours constant, whereas MART and other Methods retain the full neighbourhood graph. Additionally, not all pedestrian nodes affect each other in this scene, which has two independent pedestrian paths.

In terms of accuracy, Duri Morning yields the worst trajectory prediction performance. We used the same trajectory horizon for learning and prediction across all scenes. This means that the pedestrians in Duri Morning scenes that have the highest average velocity (see Table \ref{tab:dataset_pattern}), travel more than other scenes, meaning pedestrians travel farther within the same prediction horizon.

On the MART-SMILE methods, we applied the SMILETrack output to trajectory prediction. We observe that the training time fluctuates relative to the MART with ground truth training methods. On Duri Morning and Duri Platform 1, it is bigger, and on the other scene, it is smaller. This happened due to noise from tracking input, such as false negatives or positives. However, the tracking accuracy does not seem to be reduced significantly even when the training time is reduced significantly. This can indicate an opportunity for methods to simplify dense crowd computation to be more efficient. 

We also extend the experiments on the two best algorithms, which are SocialVAE* and MART, to work on real-world trajectories. Our SocialVAE* dominates in terms of accuracy and generalization ability across most of the scenes compared to MART. This occurs because MART considers all agents jointly, which leads to scene-specific overfitting, whereas SocialVAE* only considers a limited neighbourhood, which means it has local context for the data, making it more generalizable. However, since crowd behavior is inherently dynamic, both algorithms still require substantial training time, highlighting an open challenge in building adaptability in an efficient way.

\section{Conclusion and Future Research}
Trajectory prediction in dense, occluded crowds, which is common in real-world CCTV deployments, remains poorly evaluated by existing benchmarks, which largely feature sparse and unoccluded scenes. We introduced CrowdTraj, a dataset of 5 scenes capturing real-world CCTV footage with which reach peak densities of up to 372 pedestrians per frame, enabling end-to-end evaluation across detection, tracking, and trajectory prediction under realistic occlusion and noise.

 The current tracking methods struggle with accuracy under heavy occlusion in a real-time manner. On the other hand, the existing trajectory prediction methods struggle more with computational effort. MART achieved the best accuracy and fastest training time on pixel-coordinate prediction, while our modified SocialVAE generalized better on real-world coordinates, suggesting that local context plays a key role in scaling trajectory prediction to dense crowds. Therefore, training efficiency in dynamic, high-density scenarios remains an open challenge.

 \section{Acknowledgement}
This research was funded by the Center of Higher Education Funding and Assessment (PPAPT), the Indonesian Ministry of Higher Education and Research, the Indonesian Education Scholarship (BPI).

\bibliography{aaai25}

@article{XuVAE2022,
   author = {Pei Xu and Jean-Bernard Hayet and Ioannis Karamouzas},
   doi = {10.1007/978-3-031-19772-7_30},
   pages = {511-528},
   title = {SocialVAE: Human Trajectory Prediction Using Timewise Latents},
   url = {https://link.springer.com/10.1007/978-3-031-19772-7-30},
   year = {2022},
}

@manual{yolo2023,
  author = {Jocher, Glenn and Chaurasia, Ayush and Qiu, Jing},
  title         = "YOLO by Ultralytics",
  organization  = "Ultralytics, Inc.",
  address       = "{https://github.com/ultralytics/ultralytics}",
  year          = "2023",
  note          = "Version 8.0.0"
}

@booksection{Zhang2022,
   author = {Yifu Zhang and Peize Sun and Yi Jiang and Dongdong Yu and Fucheng Weng and Zehuan Yuan and Ping Luo and Wenyu Liu and Xinggang Wang},
   doi = {10.1007/978-3-031-20047-2_1},
   month = {10},
   pages = {1-21},
   title = {ByteTrack: Multi-object Tracking by Associating Every Detection Box},
   url = {https://link.springer.com/10.1007/978-3-031-20047-2-1},
   year = {2022},
}

@inproceedings{Aharon2022,
   author = {Nir Aharon and Roy Orfaig and Ben-Zion Bobrovsky},
   month = {6},
   title = {BoT-SORT: Robust Associations Multi-Pedestrian Tracking},
   url = {http://arxiv.org/abs/2206.14651},
   year = {2022},
}

@inproceedings{Ivanovic2019,
   author = {Boris Ivanovic and Marco Pavone},
   doi = {10.1109/ICCV.2019.00246},
   isbn = {978-1-7281-4803-8},
   journal = {2019 IEEE/CVF International Conference on Computer Vision (ICCV)},
   month = {10},
   pages = {2375-2384},
   publisher = {IEEE},
   title = {The Trajectron: Probabilistic Multi-Agent Trajectory Modeling With Dynamic Spatiotemporal Graphs},
   url = {https://ieeexplore.ieee.org/document/9009454/},
   year = {2019},
}

@inbook{Salzmann2020,
   author = {Tim Salzmann and Boris Ivanovic and Punarjay Chakravarty and Marco Pavone},
   doi = {10.1007/978-3-030-58523-5_40},
   pages = {683-700},
   title = {Trajectron++: Dynamically-Feasible Trajectory Forecasting with Heterogeneous Data},
   url = {https://link.springer.com/10.1007/978-3-030-58523-5-40},
   year = {2020},
}

@inproceedings{Zhou2022,
   author = {Rui Zhou and Hongyu Zhou and Huidong Gao and Masayoshi Tomizuka and Jiachen Li and Zhuo Xu},
   doi = {10.1109/ICRA46639.2022.9811585},
   isbn = {9781728196817},
   issn = {10504729},
   journal = {Proceedings - IEEE International Conference on Robotics and Automation},
   pages = {805-811},
   publisher = {Institute of Electrical and Electronics Engineers Inc.},
   title = {Grouptron: Dynamic Multi-Scale Graph Convolutional Networks for Group-Aware Dense Crowd Trajectory Forecasting},
   year = {2022},
}

@inproceedings{Alahi2016,
   author = {Alexandre Alahi and Kratarth Goel and Vignesh Ramanathan and Alexandre Robicquet and Li Fei-Fei and Silvio Savarese},
   doi = {10.1109/CVPR.2016.110},
   isbn = {978-1-4673-8851-1},
   booktitle = {2016 IEEE Conference on Computer Vision and Pattern Recognition (CVPR)},
   month = {6},
   pages = {961-971},
   publisher = {IEEE},
   title = {Social LSTM: Human Trajectory Prediction in Crowded Spaces},
   url = {http://ieeexplore.ieee.org/document/7780479/},
   year = {2016},
}

@article{Rudenko2020,
   author = {Andrey Rudenko and Luigi Palmieri and Michael Herman and Kris M. Kitani and Dariu M. Gavrila and Kai O. Arras},
   doi = {10.1177/0278364920917446},
   issn = {17413176},
   issue = {8},
   journal = {International Journal of Robotics Research},
   month = {7},
   pages = {895-935},
   publisher = {SAGE Publications Inc.},
   title = {Human motion trajectory prediction: a survey},
   volume = {39},
   year = {2020},
}

@inproceedings{Jiang2023,
   author = {Baowei Jiang and Bing Bai and Haozhe Lin and Yu Wang and Yuchen Guo and Lu Fang},
   doi = {10.1109/CVPR52729.2023.01581},
   isbn = {979-8-3503-0129-8},
   booktitle = {2023 IEEE/CVF Conference on Computer Vision and Pattern Recognition (CVPR)},
   month = {6},
   pages = {16479-16488},
   publisher = {IEEE},
   title = {DartBlur: Privacy Preservation with Detection Artifact Suppression},
   url = {https://ieeexplore.ieee.org/document/10204654/},
   year = {2023}
}

@inproceedings{Sundararaman2021,
   author = {Ramana Sundararaman and Cedric De Almeida Braga and Eric Marchand and Julien Pettre},
   doi = {10.1109/CVPR46437.2021.00386},
   isbn = {978-1-6654-4509-2},
   journal = {2021 IEEE/CVF Conference on Computer Vision and Pattern Recognition (CVPR)},
   month = {6},
   pages = {3864-3874},
   publisher = {IEEE},
   title = {Tracking Pedestrian Heads in Dense Crowd},
   url = {https://ieeexplore.ieee.org/document/9577483/},
   year = {2021},
}

@inproceedings{Xugroupnet2022,
   
   author = {Chenxin Xu and Maosen Li and Zhenyang Ni and Ya Zhang and Siheng Chen},
   doi = {10.1109/CVPR52688.2022.00639},
   isbn = {978-1-6654-6946-3},
   booktitle = {2022 IEEE/CVF Conference on Computer Vision and Pattern Recognition (CVPR)},
   month = {6},
   pages = {6488-6497},
   publisher = {IEEE},
   title = {GroupNet: Multiscale Hypergraph Neural Networks for Trajectory Prediction with Relational Reasoning},
   url = {https://ieeexplore.ieee.org/document/9879068/},
   year = {2022}
}

@inproceedings{Lee2024,
   author = {Seongju Lee and Junseok Lee and Yeonguk Yu and Taeri Kim and Kyoobin Lee},
   doi = {10.1007/978-3-031-72848-8_6},
   booktitle = {ECCV(European Conference of Computer Vision)},
   pages = {89-107},
   publisher = {ECVA(Europen Computer Vision Association)},
   title = {MART: MultiscAle Relational Transformer Networks for Multi-agent Trajectory Prediction},
   url = {https://link.springer.com/10.1007/978-3-031-72848-8_6},
   year = {2024}
}

@inproceedings{Cao2023,
   author = {Jinkun Cao and Jiangmiao Pang and Xinshuo Weng and Rawal Khirodkar and Kris Kitani},
   doi = {10.1109/CVPR52729.2023.00934},
   isbn = {979-8-3503-0129-8},
   booktitle = {2023 IEEE/CVF Conference on Computer Vision and Pattern Recognition (CVPR)},
   month = {6},
   pages = {9686-9696},
   publisher = {IEEE},
   title = {Observation-Centric SORT: Rethinking SORT for Robust Multi-Object Tracking},
   url = {https://ieeexplore.ieee.org/document/10204818/},
   year = {2023}
}

@inproceedings{chib2024,
author = {Chib, Pranav Singh and Singh, Pravendra},
title = {Pedestrian trajectory prediction with missing data: datasets, imputation, and benchmarking},
year = {2024},
isbn = {9798331314385},
publisher = {Curran Associates Inc.},
address = {Red Hook, NY, USA},
articleno = {3956},
numpages = {17},
location = {Vancouver, BC, Canada},
series = {NIPS '24}
}

@article{Sun2025,
   author = {Kailai Sun and Xinwei Wang and Shaobo Liu and Qianchuan Zhao and Gao Huang and Chang Liu},
   doi = {10.34740/kaggle/ds/7494891},
   title = {Towards pedestrian head tracking: A benchmark dataset and a multi-source data fusion network},
   year = {2025}
}

@article{Dufour2025,
   author = {Oscar Dufour and Huu Tu Dang and Jakob Cordes and Raphael Korbmacher and Mohcine Chraibi and Benoit Gaudou and Alexandre Nicolas and Antoine Tordeux},
   doi = {10.1038/s41597-025-04732-3},
   issn = {20524463},
   issue = {1},
   journal = {Scientific Data },
   month = {12},
   pmid = {40307272},
   publisher = {Nature Research},
   title = {Dense Crowd Dynamics and Pedestrian Trajectories: A Multiscale Field Dataset from the Festival of Lights in Lyon},
   volume = {12},
   year = {2025}
}

@inproceedings{Lin2024,
   author = {Haozhe Lin and Chunyu Wei and Li He and Yuchen Guo and Yunqi Zhao and Shanglong Li and Lu Fang},
   doi = {10.1109/CVPR52733.2024.01829},
   isbn = {9798350353006},
   issn = {10636919},
   booktitle = {Proceedings of the IEEE Computer Society Conference on Computer Vision and Pattern Recognition},
   pages = {19331-19340},
   publisher = {IEEE Computer Society},
   title = {GigaTraj: Predicting Long-term Trajectories of Hundreds of Pedestrians in Gigapixel Complex Scenes},
   year = {2024}
}

@inproceedings{WongSC2024,
   author = {Conghao Wong and Beihao Xia and Ziqian Zou and Yulong Wang and Xinge You},
   booktitle = {2023 IEEE/CVF Conference on Computer Vision and Pattern Recognition (CVPR)},
   pages = {19005-19015},
   title = {SocialCircle: Learning the Angle-based Social Interaction Representation for Pedestrian Trajectory Prediction},
   url = {https://github.},
   year = {2024}
}

@article{Cao2017,
   author = {Shuchao Cao and Armin Seyfried and Jun Zhang and Stefan Holl and Weiguo Song},
   doi = {10.1088/1742-5468/aa620d},
   issn = {17425468},
   issue = {3},
   journal = {Journal of Statistical Mechanics: Theory and Experiment},
   month = {3},
   publisher = {Institute of Physics Publishing},
   title = {Fundamental diagrams for multidirectional pedestrian flows},
   volume = {2017},
   year = {2017}
}

@inproceedings{Pellegrini2009,
   author = {S Pellegrini and A Ess and K Schindler and L van Gool},
   doi = {10.1109/ICCV.2009.5459260},
   isbn = {978-1-4244-4420-5},
   booktitle = {2009 IEEE 12th International Conference on Computer Vision},
   month = {9},
   pages = {261-268},
   publisher = {IEEE},
   title = {You'll never walk alone: Modeling social behavior for multi-target tracking},
   url = {http://ieeexplore.ieee.org/document/5459260/},
   year = {2009}
}

@article{Lerner2007,
   author = {Alon Lerner and Yiorgos Chrysanthou and Dani Lischinski},
   doi = {10.1111/j.1467-8659.2007.01089.x},
   issn = {14678659},
   issue = {3},
   journal = {Computer Graphics Forum},
   pages = {655-664},
   publisher = {Blackwell Publishing Ltd},
   title = {Crowds by example},
   volume = {26},
   year = {2007}
}

@inbook{Robicquet2016,
   author = {Alexandre Robicquet and Amir Sadeghian and Alexandre Alahi and Silvio Savarese},
   doi = {10.1007/978-3-319-46484-8_33},
   pages = {549-565},
   title = {Learning Social Etiquette: Human Trajectory Understanding In Crowded Scenes},
   url = {http://link.springer.com/10.1007/978-3-319-46484-8_33},
   year = {2016}
}

@inproceedings{Fu2025,
   author = {Yuxiang Fu and Qi Yan and Lele Wang and Ke Li and Renjie Liao},
   booktitle = {2025 IEEE/CVF Conference on Computer Vision and Pattern Recognition (CVPR)},
   pages = {17282-17293},
   publisher = {theCVF},
   title = {MoFlow: One-Step Flow Matching for Human Trajectory Forecasting via Implicit Maximum Likelihood Estimation based Distillation},
   year = {2025}
}

@software{yolo11_ultralytics,
  author = {Glenn Jocher and Jing Qiu},
  title = {Ultralytics YOLO11},
  version = {11.0.0},
  year = {2024},
  url = {https://github.com/ultralytics/ultralytics},
  orcid = {0000-0001-5950-6979, 0000-0003-3783-7069},
  license = {AGPL-3.0}
}

@software{yolo26_ultralytics,
  author = {Glenn Jocher and Jing Qiu},
  title = {Ultralytics YOLO26},
  version = {26.0.0},
  year = {2026},
  url = {https://github.com/ultralytics/ultralytics},
  orcid = {0000-0001-5950-6979, 0000-0003-3783-7069},
  license = {AGPL-3.0}
}

@article{WangSmile2024,
   author = {Yu-Hsiang Wang and Jun-Wei Hsieh and Ping-Yang Chen and Ming-Ching Chang and Hung-Hin So and Xin Li},
   doi = {10.1609/aaai.v38i6.28386},
   issn = {2374-3468},
   issue = {6},
   journal = {Proceedings of the AAAI Conference on Artificial Intelligence},
   month = {3},
   pages = {5740-5748},
   publisher = {TheCVF},
   title = {SMILEtrack: SiMIlarity LEarning for Occlusion-Aware Multiple Object Tracking},
   volume = {38},
   url = {https://ojs.aaai.org/index.php/AAAI/article/view/28386},
   year = {2024}
}

@article{Chen2025,
   author = {Wangxing Chen and Haifeng Sang and Jinyu Wang and Zishan Zhao},
   doi = {10.1109/TITS.2024.3525080},
   issn = {15580016},
   issue = {5},
   journal = {IEEE Transactions on Intelligent Transportation Systems},
   pages = {6923-6935},
   publisher = {Institute of Electrical and Electronics Engineers Inc.},
   title = {DSTIGCN: Deformable Spatial-Temporal Interaction Graph Convolution Network for Pedestrian Trajectory Prediction},
   volume = {26},
   year = {2025}
}

@inproceedings{Erregue2025,
   author = {Iñaki Erregue and Kamal Nasrollahi and Sergio Escalera},
   booktitle = {2025 IEEE/CVF Winter Conference on Applications of Computer Vision Workshops (WACVW)},
   pages = {824-833},
   publisher = {theCVF},
   title = {YOLO11-JDE: Fast and Accurate Multi-Object Tracking with Self-Supervised Re-ID},
   url = {https://github.com/},
   year = {2025}
}

@article{Badauraudine2025,
   author = {Muhammad Firdaus Mohamed Badauraudine and Megat Norulazmi Megat Mohamed Noor and Mohd Shahizan Othman and Haidawati Binti Mohamad Nasir},
   doi = {10.5530/irc.1.2.10},
   issue = {2},
   journal = {Information Research Communications},
   month = {1},
   pages = {65-73},
   publisher = {Manuscript Technomedia LLP},
   title = {Detection and Tracking of People in a Dense Crowd through Deep Learning Approach-A Systematic Literature Review},
   volume = {1},
   year = {2025}
}

\end{document}